\documentclass[sigconf]{acmart}

\AtBeginDocument{%
  }

\copyrightyear{2024}
\acmYear{2024}
\setcopyright{rightsretained}
\acmConference[MMSports '24]{Proceedings of the 7th ACM International Workshop on Multimedia Content Analysis in Sports}{October 28-November 1, 2024}{Melbourne, VIC, Australia}
\acmBooktitle{Proceedings of the 7th ACM International Workshop on Multimedia Content Analysis in Sports (MMSports '24), October 28-November 1, 2024, Melbourne, VIC, Australia}
\acmDOI{10.1145/3689061.3689074}
\acmISBN{979-8-4007-1198-5/24/10}
\makeatletter
\gdef\@copyrightpermission{
  \begin{minipage}{0.3\columnwidth}
   \href{https://creativecommons.org/licenses/by-nc/4.0/}{\includegraphics[width=0.90\textwidth]{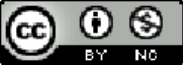}}
  \end{minipage}\hfill
  \begin{minipage}{0.7\columnwidth}
   \href{https://creativecommons.org/licenses/by-nc/4.0/}{This work is licensed under a Creative Commons Attribution-NonCommercial International 4.0 License.}
  \end{minipage}
  \vspace{5pt}
}
\makeatother
\usepackage{subcaption}
\usepackage{booktabs}

\begin{document}

\title{Enhancing Soccer Camera Calibration Through Keypoint Exploitation}

\author{Nikolay S. Falaleev}
\email{nikolay.falaleev@sportlight.ai}
\orcid{0000-0002-9398-5261}
\affiliation{%
  \institution{Sportlight}
  \city{Bicester}
  \country{United Kingdom}
}

\author{Ruilong Chen}
\email{ruilong.chen@sportlight.ai}
\orcid{0000-0002-7855-4711}
\affiliation{%
  \institution{Sportlight}
  \city{Bicester}
  \country{United Kingdom}
}

\renewcommand{\shortauthors}{Nikolay S. Falaleev and Ruilong Chen}

\begin{abstract}

Accurate camera calibration is essential for transforming 2D images from camera sensors into 3D world coordinates, enabling precise scene geometry interpretation and supporting sports analytics tasks such as player tracking, offside detection, and performance analysis. However, obtaining a sufficient number of high-quality point pairs remains a significant challenge for both traditional and deep learning-based calibration methods.
This paper introduces a multi-stage pipeline that addresses this challenge by leveraging the structural features of the football pitch. Our approach significantly increases the number of usable points for calibration by exploiting line-line and line-conic intersections, points on the conics, and other geometric features.
To mitigate the impact of imperfect annotations, we employ data fitting techniques. Our pipeline utilizes deep learning for keypoint and line detection and incorporates geometric constraints based on real-world pitch dimensions. A voter algorithm iteratively selects the most reliable keypoints, further enhancing calibration accuracy.
We evaluated our approach on the largest football broadcast camera calibration dataset available, and secured the top position in the SoccerNet Camera Calibration Challenge 2023 \cite{cioppa2023soccernet}, which demonstrates the effectiveness of our method in real-world scenarios. The project code is available at \href{https://github.com/NikolasEnt/soccernet-calibration-sportlight}{https://github.com/NikolasEnt/soccernet-calibration-sportlight}.

\end{abstract}

\ccsdesc[300]{Computing methodologies~Interest point and salient region detections}
\ccsdesc[500]{Computing methodologies~Scene understanding}
\ccsdesc[500]{Computing methodologies~Object detection}
\ccsdesc[300]{Computing methodologies~Shape inference}

\keywords{Camera Calibration, Computer Vision, Keypoint Detection, Line Detection, Football, Broadcast Video}

\begin{teaserfigure}
    \includegraphics[width=\textwidth]{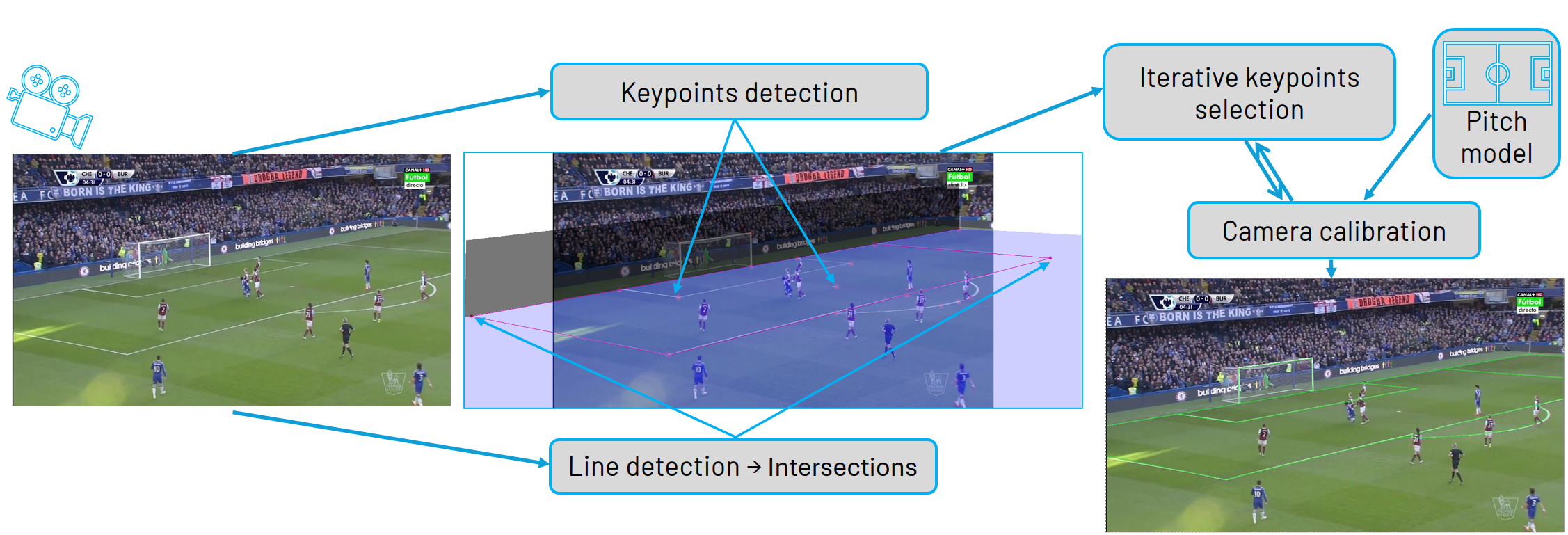}
    \caption{Proposed camera calibration pipeline applied to an image from SoccerNet-v3 dataset \cite{soccernetv3}.}
    \Description{Camera calibration stages: a video frame is processed by two models: kaypoint detection and line detection. The results are overlayed over the original image. Then, the prediction results, including lines intersection points beyond the frame boundaries, are processed by an Iterative keypoint selection algorithm, which selects points for camera calibration. Camera calibration takes the points and predefined pitch model to output camera parameters. Finally, the football pitch is overlayed over the  frame.}
\end{teaserfigure}

\maketitle

\section{Introduction}
\label{sec:intro}

Camera calibration links the recorded or broadcasted videos to the real physical world by determining the parameters of a camera. These parameters are essential for mapping 2D images captured by the camera to 3D world coordinates, enabling accurate interpretation of the scene geometry.  Camera calibration involves estimating intrinsic and extrinsic parameters from images and is crucial in contexts such as soccer games. For example, a soccer player in a video could be projected onto their location on the pitch \cite{cioppa2021camera}, aiding in the decision making processes such as offside detection or goal scoring \cite{cioppa2021camera, giancola2022soccernet, theiner2023tvcalib, cuevas2020automatic}. It also assists in the metric evaluation of player detection and tracking \cite{manafifard2024review} and provides match statistics, such as the distance covered by players and their speeds \cite{cuevas2020automatic, barros2007analysis}. Vision-based tracking systems, employed by professional sports analysis companies, heavily rely on accurate camera calibration. For example, Sportlight, which provides soccer analysis services to the English Premier League teams to help coaches improve training strategies and enhance game performance, considers its camera calibration to be one of the cornerstones for guaranteeing the accuracy of video-based data \cite{sportlight_ai}.

In a pinhole camera model, a point from 3D world coordinates maps to 2D image coordinates using a 3x3 homography matrix containing 8 independent parameters \cite{bu2011automatic}. The matrix can be calculated using at least 4 point pairs where no three of them are collinear from the model and image plane \cite{manafifard2024review}. Since there is a distinct colour/luminance difference between markings and grass, it is straightforward to utilise markings that contain position information from a soccer pitch, including the midfield line, penalty lines and circles \cite{manafifard2024review, magera2024universal}.

Despite the importance of camera calibration in soccer, several challenges persist in achieving accurate and robust calibration. One of the primary difficulties lies in detecting a sufficient number of accurate point pairs that are used for estimating the homography matrix. Traditional methods relying on low-level pixel information are often vulnerable to noise, shadows, lighting variations and occlusions \cite{manafifard2024review, bu2011automatic, homayounfar2017sports}. While recent Deep Learning approaches have shown promising results in extracting more robust features, they still face limitations in terms of generalization and the ability to fully exploit the structural information present in the soccer pitch.

These challenges have motivated researchers to explore alternative approaches that can overcome the limitations of existing methods and improve the accuracy of camera calibration in soccer. For example, some approaches directly learn the homography matrix from big amount of data, thus eliminating the step of finding the best matching points \cite{theiner2023tvcalib, shi2022self}. Meanwhile, we believe that all the current point matching approaches have not exploited structural information to its maximum potential. Hence, we propose a heatmap-based approach that finds the extremities and employs special points on the circles. This approach boosts the number of matching point pairs. Figure \ref{fig:pitch_pattern} shows 57 points we used in our point detection model, while our line detection model provides 23 lines that can further increase the number of intersections we can find and support the mis-detections from the point detection model. In addition, another limitation of the other approaches is treating annotations as the absolute truth, while we derive our keypoints using line fitting, which helps reduce noise in annotations and hence improves generalization ability.

The key contributions of this work can be summarized as follows:
\begin{itemize}
    \item \textbf{Addressing point-pair scarcity}: We propose a novel method to overcome the challenge of finding a sufficient number of point pairs for calibration within individual video frames from football broadcasts. This method leverages the structural features of the football pitch, including intersections, conics and tangency points.

    \item \textbf{Robust data processing}: We utilise computed keypoints instead of relying directly on annotation data, thus enhancing robustness against imperfections in the data. Furthermore, our camera calibration method employs a heuristic selection of predicted elements for calibration, improving the reliability of predictions even in cases of imperfect or incomplete outputs of Deep Learning models.

    \item \textbf{Multi-stage calibration pipeline}: We introduce a multi-stage pipeline for camera calibration that integrates keypoint and line detection models. This approach generates additional intersection points, leading to more accurate calibration. Our method achieved the best results on the largest camera calibration dataset, winning the Soccernet Camera Calibration Challenge 2023  \cite{cioppa2023soccernet}.

\end{itemize}

\section{Related Work}
\label{sec:related}
Camera calibration techniques in sports analysis, particularly in soccer, have evolved significantly over the years. This section provides an overview of the various approaches, from traditional to more recent deep learning based techniques. 

\subsection{Traditional Methods}
Early approaches to camera calibration in soccer heavily relied on computer vision techniques that extract features from low-level pixels. These methods, while foundational, face challenges in real-world scenarios.

The Scale Invariant Feature Transform (SIFT) \cite{sift} has been widely used for keypoint detection \cite{goorts2014self}. SIFT's ability to identify distinctive invariant features makes it suitable for matching different viewpoints of a pitch point, for example, a line intersection.  The Hough transformation has been widely used for identifying straight lines in images, which is particularly useful for detecting marking lines on the pitch \cite{niu2012tactic, manafifard2024review}. Advanced Least Square Fitting (ALSF) \cite{wang2006fast} has been applied for detection arcs and fitting ellipses, and is used for identifying pitch elements like the circles and arcs \cite{manafifard2024review}. While these methods laid the groundwork for camera calibration in sports, they are inherently limited by their reliance on low-level features. This makes them vulnerable to environmental factors such as varying lighting conditions, shadows, and occlusion, which are common in sports broadcasts \cite{manafifard2024review, bu2011automatic, homayounfar2017sports}.

\subsection{Deep Learning-based Methods}
Deep learning brings a new era of camera calibration techniques, offering more robust feature extraction and improved performance. 
Homayounfar \cite{homayounfar2017sports} introduced a semantic segmentation approach that classifies pixels into one of the six categories: vertical lines, horizontal lines, side circles, middle circle, grass and crowd. This method provides a more comprehensive understanding of the pitch layout compared to traditional line and ellipse detection. Citraro et al. \cite{citraro2020real} expanded beyond pitch markings to include centers of players' mass as additional points. While innovative, this approach requires multiple camera setups to accurately map these points to the ground plane, limiting its applicability in single camera scenarios. Moreover, the complex setup and calibration process required for multiple cameras further restricts its practical implementation in many real-world broadcasting situations. The number of point pairs is crucial in camera calibration, as fitting algorithms can be applied to reduce projection errors. To address the challenge of limited visible pitch markings, some researchers have explored estimating vanishing points outside the image frame \cite{homayounfar2017sports}. However, this approach struggles with accuracy involving distorted perspectives \cite{citraro2020real}. Nie et al. \cite{nie2021robust} proposed using uniform grids across the entire field template. While this uniform distributed points could easily cause misalignment problems in predictions. Building upon the grid-based approach, Chu et al. \cite{chu2022sports} developed an instance segmentation approach that assigns separate labels to each point. This gives the network additional context, which could lead to a better model. However, since grid points do not form a line or a corner and rely heavily on the training dataset, the lack of connection with physical structure of the pitch decreases its generalization ability.

Despite these advancements in camera calibration, a significant challenge persists: obtaining a sufficient number of accurate point pairs for reliable calibration. The grid-based approach \cite{nie2021robust, chu2022sports} demonstrates that even points not on a line or circle can contribute to learning spatial relationships within the image. This highlights a crucial insight: exploiting more points from inherent structural information, rather than relying on a grid, could substantially enhance calibration accuracy. Our work hence focuses on points that have clear correspondences on structural pitch markings, aiming to improve both the quantity and quality of point pairs used in camera calibration.

\section{Method}
\label{sec:method}

\begin{figure}[t]
\begin{center}
   \includegraphics[width=1.0\linewidth]{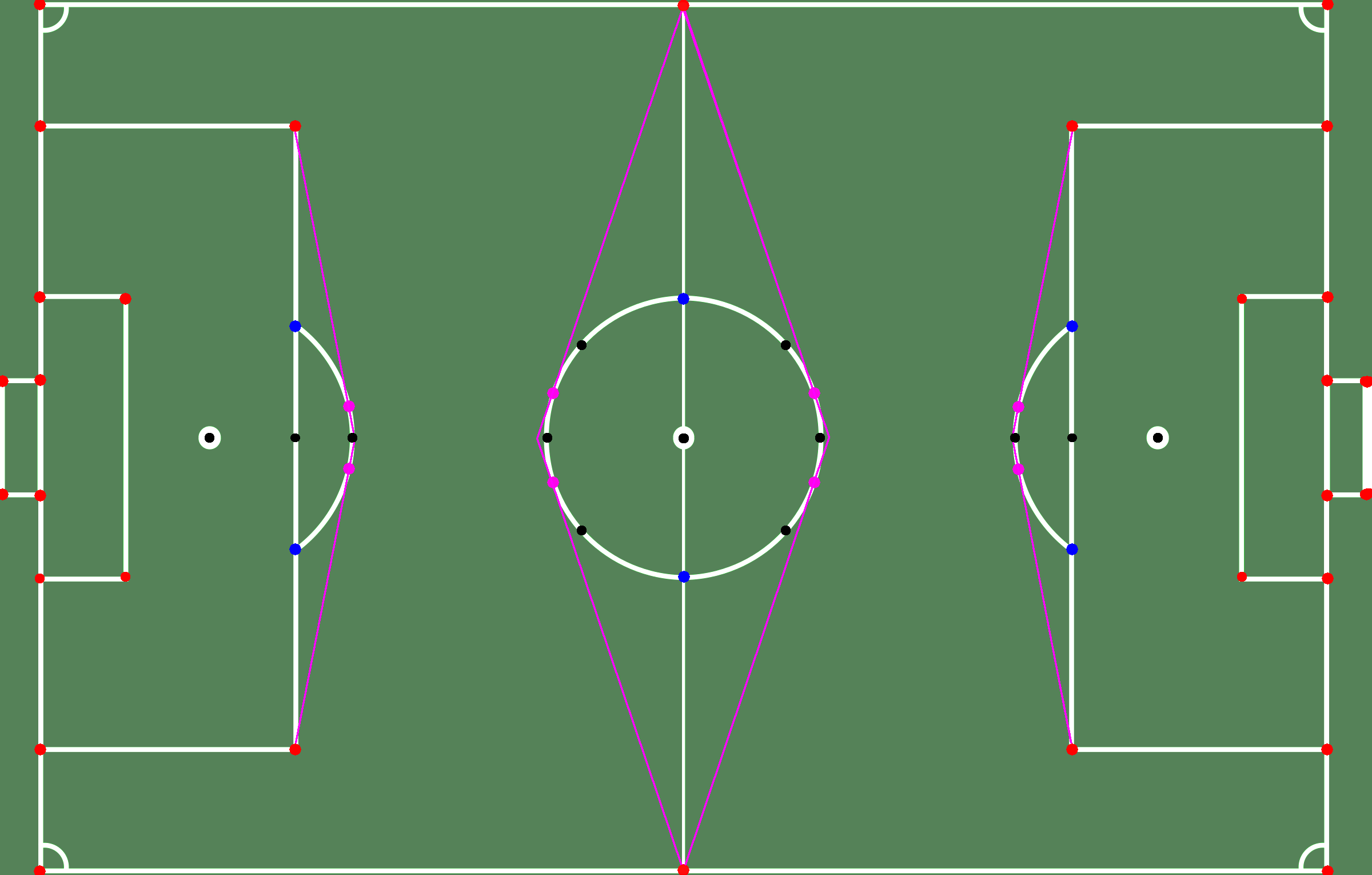}
\end{center}
   \caption{Keypoints used as the points detection model target. \emph{Red} - points of line-line intersections; \emph{Blue} - line-conic intersection points; \emph{Purple} - conic tangent points; \emph{Black} - other points projected by homography.}
   \Description{This figure depicts a football pitch pattern with keypoint locations marked. Keypoints are positioned at intersections of geometric features and in other strategic places, such as along the longitudinal line features (center of the pitch, intersection with lines and conics, penalty mark and tangent points to the circles.}
\label{fig:pitch_pattern}
\end{figure}

\subsection{Points Detection}
\begin{figure}[t]
\begin{center}
   \includegraphics[width=1\linewidth]{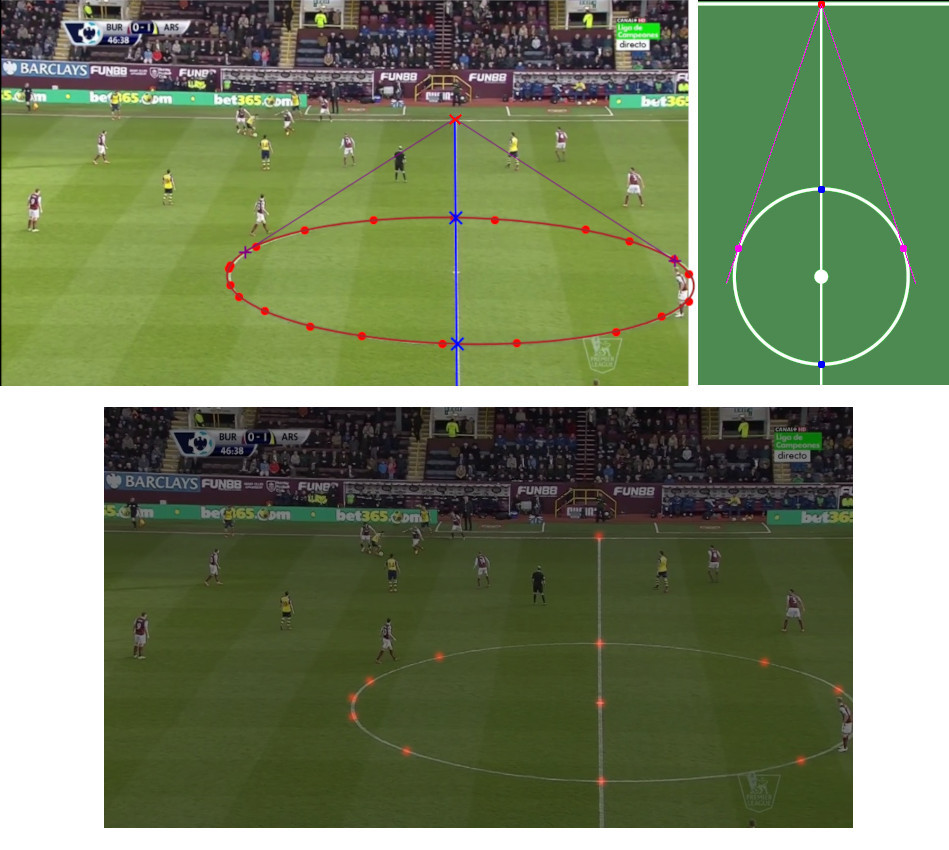}
\end{center}
   \caption{Top row (left): Red dots denote the central circle points from the annotation. The red curve represents the ellipse fit to these points. Blue crosses indicate computed line-conic intersections, the red cross marks a computed line-line intersection, and purple crosses represent the derived tangent points to the ellipse. Note that the intersection points (red and blue crosses) are insufficient for homography computation. The corresponding points on the pitch pattern are shown on the right.
   Bottom row: Heatmaps of all target points overlaid on the video frame.}
   \Description{An illustration of points derivation. The first image shows a video frame from football broadcast, where the central circle is visible. It displays overlaid points from the circle annotation and computed keypoint positions. All intersection points lie on the pitch's central line. The bottom image shows the derived heatmaps overlaid on the video frame.}
\label{fig:tangent}
\end{figure}

Points, lines, circles or ellipses are the most distinguished features on a pitch, and  are widely used for soccer pitch camera calibration, especially directly with extremities \cite{magera2024universal, giancola2022soccernet, cioppa2023soccernet}. However, there are two issues using extremities, one is the scarcity of points, especially when camera field of view does not include many of these points. The second reason is that annotating extremities precisely is challenging. In this paper, our points are mainly derived from intersections generated from lines. We exploit points from the geometrical structure of the pitch. In total, we define 57 landmark keypoints on a football pitch pattern (Fig. \ref{fig:pitch_pattern}). These points consist of:

\begin{itemize}
    \item \textbf{Line-line intersection}: Thirty points are defined as the intersections of linear data fittings applied to the line points provided in the annotation. We choose intersections of lines rather than line extremities because extremities could be missing or exhibit significant annotation errors. A two-step line-fitting based intersection point generation process is employed. Initially, we perform line fitting using all available annotation points for each line of interest. The resulting line equations enable the derivation of a coarse intersection point, as actual lines in video frames may appear curved rather than straight due to lens geometric distortion. The statistical aggregation inherent in fitting lines to the annotation points reduces the influence of individual point annotation errors on the final intersection calculation.
    In the second stage, the coarse intersection location is used to select a few annotation points for each line that are proximate to the intersection point. These points are then fitted with lines to determine the refined keypoint location. This two-step approach facilitates more accurate intersection computation and filters out some outliers.
    \item \textbf{Line-conic intersection}: Six points were defined as intersections between conics ('Circle central,' 'Circle left,' and 'Circle right' from the annotation) and lines. The conics points from annotation are fitted with ellipses using Halíř–Flusser's Least Squares ellipse fitting algorithm \cite{halir1998numerically, ben_hammel_2020_3723294}. The intersections were derived analytically as points of ellipse-line intersection, utilising the equations of the fitted ellipses and lines.
    \item \textbf{Conic tangent}: Eight points were defined as the tangent points of tangent lines from a known external point to an ellipse. In many cases, there are insufficient visible intersections to construct a homography, while circles are present. This is common, for example, in frames where only the pitch centre is visible. To address this, we utilized correspondence between tangent points of tangent lines from a known point to the circles, augmenting the available points for homography construction. Football pitch lines intersections were used as the known external points through which the tangent lines should pass (Fig. \ref{fig:tangent}).
    Real-world coordinates for the points were derived as tangent points of a line passing through the given external point and the corresponding circle. This exploits the fact that projective transformations preserve tangency. Note that the computation disregarded the effect of geometric lens distortion.
    \item \textbf{Additional structural points}: Using the homography created with points from the aforementioned subsets, we added nine points along the longitudinal pitch axis (including the pitch centre and the penalty spots), as well as four points marking quarter turns on the central circle. This approach, applying the homography to a subset of desired keypoints, utilises corresponding real-world points. It also allows for the inclusion of missing points, such as when lines are missed in the annotation.
\end{itemize}

In addition to exploiting the extracted points, we also addressed the challenge of left-right ambiguity. In scenarios where the camera alignment coincides with the direction of the pitch’s long axis, accurately differentiating between the left and right sides without temporal context becomes difficult. However, it is crucial to establish a clear distinction between the two sides in the ground-truth values for model training, particularly when both goals are visible. Without resolving this ambiguity, the model might be unsure whether a keypoint, such as the corner of a goalpost, is on the left or right side of the pitch, potentially resulting in the prediction of both left and right keypoints at the same spatial location within a frame.

We implemented a remapping process to resolve this left-right ambiguity and ensure consistency. The points are remapped so that the goal area closest to the camera is consistently considered the left side.

\subsection{Line Detection}

\begin{figure}[t]
\begin{center}
   \includegraphics[width=0.7\linewidth]{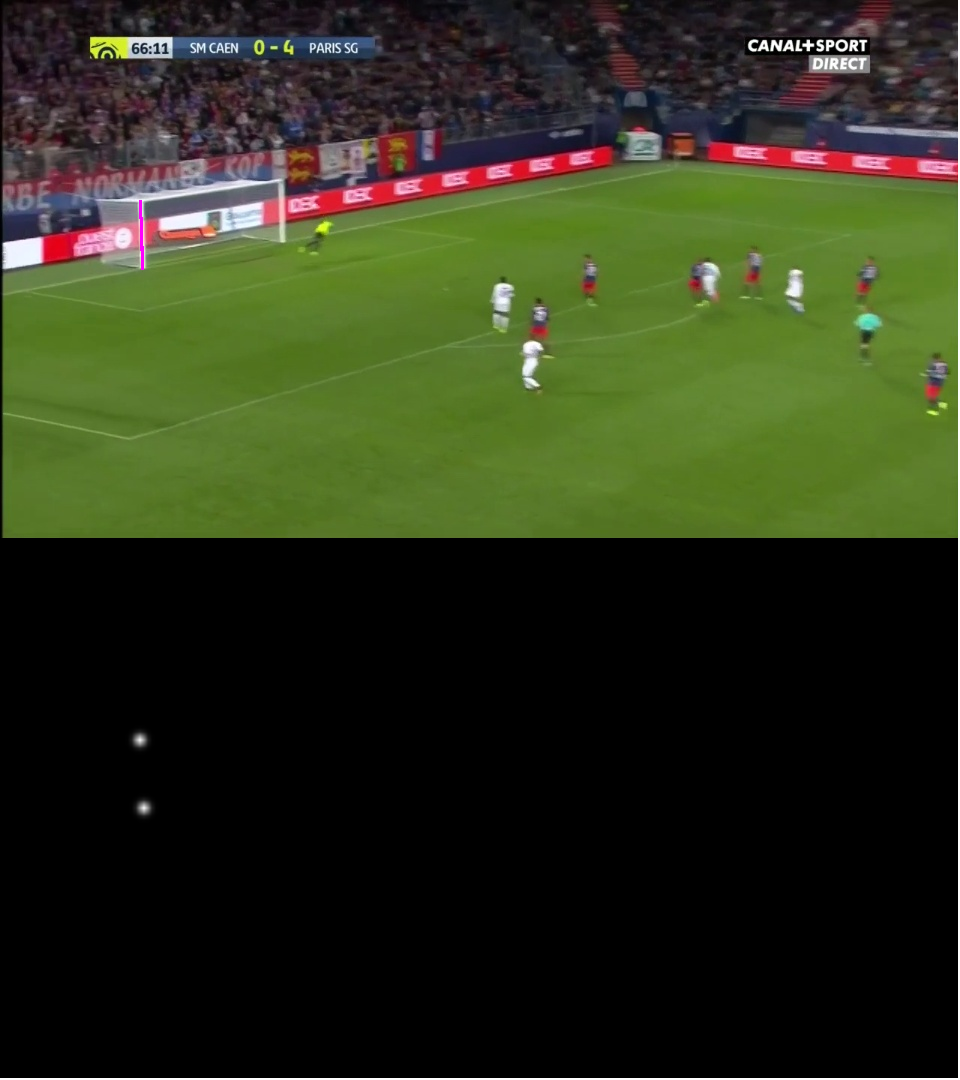}
\end{center}
   \caption{Original image with a magenta line overlay for 'Goal left post left' line (top) and the corresponding heatmap channel of the line detection target (bottom).}
   \Description{A football broadcast video frame and its corresponding heatmap for one of the lines, containing two Gaussian peaks to highlight the extremities of the line.}
\label{fig:lines_heatmap}
\end{figure}

Besides the point detection model that generates points with meaningful structural insights, we have also implemented a line detection model to aid in some challenging scenarios. This augmentation ensures sufficient points for camera calibration, when keypoints within the image are insufficient, as line intersection points beyond the image boundary are considered. In addition, some missing points from the point detection model are available in the line detection model. The line detection still uses two extremities of a line, however, it has two main advantages over directly applying extremities for camera calibration. Firstly, precise extremities are challenging to annotate. Secondly, if the predicted point is incorrect but still on the line, it does not affect the accuracy of the line.  In order to detect extremities and then fit a line, a heatmap-based model is applied similarly to the point detection model. 

To handle unsorted points in the annotations, a line fitting algorithm is applied to obtain valid extremities. Points that failed the line fitting algorithm are excluded in training. Since defining start and end points is challenging, extremities are considered orderless. The head of the network is a heatmap layer of 23 channels, where each channel corresponds to a line. For channels representing existing lines, two heat points are generated. Fig. \ref{fig:lines_heatmap} shows an example of heatmap channel labelled 'Goal left post left' for a football video frame.

\section{Experiments}
\subsection{Data and Metric}

Datasets used for camera calibration in soccer are quite limited in the literature. The WorldCup dataset \cite{homayounfar2017sports} and TS-WorldCup \cite{homayounfar2017sports} are small in terms of size, with 395 and 3,812 images, respectively. These datasets are not widely used for several reasons. Deep learning models typically require larger datasets for effective training. Additionally, the datasets contain homographies as annotations. Homography annotations have limitations because they are specific to a particular camera model and often fail to account for lens distortions, especially radial distortions, which are common in broadcast cameras. In contrast, field marking annotations recently introduced by Magera et al. \cite{magera2024universal} offer significant advantages. They label semantic points on soccerfield markings which are independent of the cameras used, meaning they are valid for any type of camera. In addition, field marking annotations facilitate more complex calibration scenarios; for example, links among points open the possibility to find occluded or out of bound. In fact, linking different points is the key to our success in this paper. There are two datasets using field marking annotations, the Soccernet-Calibration-2022 \cite{giancola2022soccernet}, consisting of 21,132 images and Soccernet-Calibration-2023 with 25,506 images in 960x540 px resolution, generated from 500 games. In this paper, we evaluate our approach on Soccernet-Calibration-2023 because it encompasses the former, with more testing images added.

Since ground truth camera parameters are not directly available in this dataset, we can evaluate the quality of the estimated camera parameters by projecting 3D field points to the image plane and then comparing the errors with manual annotations, which are also in the 2D image plane. Point location errors do not consider distortions; for example, it is possible that the endpoints of a distorted line are close, but the distance between the middle points is large. As 2D points of a field segment form polylines, we aim to overlap the predicted polyline with the annotated polylines as much as possible. Magera et al. \cite{magera2024universal} propose an advanced metric for accuracy that measures how well the two polylines match. This metric therefore considers distortion and aligns better with human inspection. For points from a ground truth field segment, if all of them have a shorter distance than a threshold ($t$ pixels) to the polyline segment formed by predicted points, this segment is considered as a true positive ($TP@t$). If any point from annotation does not meet this condition, or a segment appears in the prediction, but not in annotations, it counts as a false positive ($FP@t$). If a segment appears in annotation but not in prediction, it is considered as a false negative ($FN@t$). The accuracy metric is defined as:

\begin{equation}
    Acc@t = \frac{TP@t}{TP@t + FN@t +FP@t}.
\end{equation}
The SoccerNet Camera Calibration Challenge 2023 adopts this metric and fixes $t=5$ \cite{cioppa2023soccernet}. Additionally, a higher rate of images with predicted camera parameters is desirable. This is measured by the Completeness Ratio ($CR$). The final metric is a multiplication of the Completeness Ratio and the accuracy components:
\begin{equation}
  Score = Acc@5 \times CR
\end{equation}
It is worth noting that the CR metric, without the context of accuracy results, provides little insight into the actual performance of an approach, as the value can reach its maximum of 1.0, for instance, when all images are provided with default camera parameters.

In addition, to evaluate the quality of the raw predicted keypoints, we adopt the $L_{2}$ metric, which assesses the average $L_{2}$ distance between corresponding points in the generated ground truth and predicted keypoint locations.

\begin{figure}[h]
  \centering
  \includegraphics[width=\linewidth]{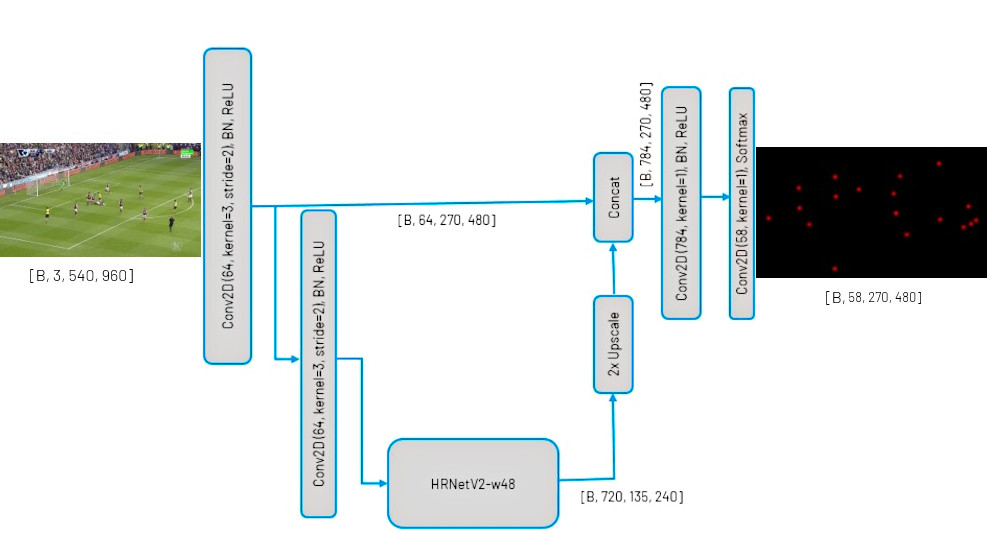}
   \caption{The keypoint detection model architecture. The line model follows the same structure but differs in the number of target featuremaps.}
   \Description{This model architecture based on the HRNetV2-w48 backbone. The input image is first processed by a Conv2D layer with a stride of 2, reducing the spatial resolution. The resulting tensor is then passed through the backbone, producing an output with a resolution of 1/4 the original image size. This output tensor is upscaled by a factor of 2 and concatenated with the tensor from before the backbone. A final Conv2D layer is applied to produce the required number of channels.}
\label{fig:architecture}
\end{figure}

\subsection{Neural Network Models}
Both the point detection model and the line model are based on keypoint extraction using heatmaps. Since the focus of this paper is on point exploitation rather than comparisons of different semantic models, we employed the HRNetV2-w48 backbone \cite{wang2020deep}, which is commonly used for keypoint extraction \cite{wakai2024deep}. Our results demonstrate that the proposed point exploitation method achieves the best performance on the largest publicly available camera calibration dataset. Using more advanced keypoint detection models and backbones could further improve camera calibration accuracy. However, exploring alternative models is beyond the scope of this work. The two keypoint models were trained with different settings and target feature maps.

\subsubsection{Point Detection}

To enhance spatial resolution of the predicted heatmaps in point detection model, we added 2x upsampling and skip-connection features concatenation from the corresponding resolution of the convolution stem. The final predictions had half the resolution of the original image, with Softmax used as the last activation. The target tensor consisted of 2D feature maps for each point, where Gaussian peaks (\( \sigma = 3 px \)) were positioned at the keypoint locations. An extra target channel was included, representing the inverse of the maximal value among the other target feature maps. This ensured that the final target tensor summed to 1.0 at each spatial point. The location of the maximal value in the predicted heatmaps for each class was considered the predicted keypoint position, adjusted for the 50\% downscaling of the predictions spatial resolution. The value itself was considered the keypoint prediction confidence. Only points within the image boundaries were used as targets. The details of the architecture are depicted in Fig. \ref{fig:architecture}.

The model was trained using MSE loss and the Adam optimizer \cite{kingma2014adam}. The initial learning rate was 0.001 and was halved when the loss did not improve for 8 consecutive epochs. This halving continued until there was no improvement for 32 epochs. Subsequently, the neural network was finetuned using the Adaptive Wing Loss (AWL) \cite{wang2019adaptive} with the same strategy and starting learning rate of 5e-4. The best checkpoint was selected based on the combined metric value on the validation dataset.

During training, the data was augmented by horizontal flipping, Gaussian noise, and random colour variations (including image brightness, colour and contrast). Each transformation was applied with a probability of 0.5. The backbones were initialized with open-sourced ImageNet pretrained weights.

On the inference stage, using a single Nvidia GeForce RTX 3090 GPU, the architecture achieves a performance rate of 44.1 ms per image for batches consisting of a single image and 3.1 ms for batches of 128 images, without any specific hardware-related optimisation.

Although real-time performance is beyond the scope of this work, the results suggest that the proposed architecture could serve as a basis for further development of a real-time solution.

\subsubsection{Line Detection}
The line detection utilized the same model backbone as in the point detection model with a heatmaps prediction size one-quarter the resolution of the original image. During the initial training process, extremities with their 8 neighbourhood locations on heatmaps were set to peaks, followed by a Gaussian blur being applied on each heatmap channel. AWL and the Adam optimizer ($lr = 0.001$) was applied. Subsequently, extremities were changed to Gaussian peaks (\( \sigma = 4 px \)) as shown in Fig. \ref{fig:lines_heatmap}, the network was then fined tuned using a summed loss adding the Gaussian MSE with AWL.

In addition, in order to guarantee the quality of the ground truth, points failed in the line fitting algorithm were not be used for training and validation. During the training, various data augmentations were triggered at a certain probability, including colour augmentation, Gaussian noise and horizontal flip. To decode the heatmaps back into lines, a process involved finding the first point with the highest confidence, applying a Gaussian mask to this point area, locating the second peak, and ultimately producing the line equation.

In terms of speed, the line model required an average processing time of 33.6 ms per image on a single Nvidia GeForce RTX 3090 GPU with a batch size of 1.

\subsection{Camera Calibration}
This section describes several steps of increasing complexity in the camera calibration process. We used the base camera calibration algorithm implemented in the calibrateCamera method from OpenCV 4.7 \cite{opencv_library} (the algorithm is referred to as OpenCV hereafter). A standard pinhole camera model was utilized, assuming zero astigmatism and distortions, and fixing the principal point to the center of the frame. Keypoints, along with the actual pitch pattern dimensions, were used instead of a conventional checkerboard calibration pattern.
Despite attempts to optimize the lens distortion parameters, the results were of lower accuracy, possibly due to imperfections in the annotation and insufficient accuracy of the predictions.

Improvements to the basic calibration approach included considering two additional vertical planes containing the goal polygons. This leveraged the fact that the football pitch pattern forms a 3D grid, given that the crossbars are off the ground plane (referred to as OpenCV Multiplane). Effectively, this approach uses up to three planes for calibration, similar to traditional checkerboard calibration using up to three images. Two of these planes are highlighted with grey and blue shades in the central section of Fig. 1.

In some cases where intersections were sufficient for reliable calibration, we found that adding extra points actually decreased calibration accuracy. Furthermore, in some rare cases, some predicted points were outliers. Filtering such points had the potential to improve the accuracy of the camera parameters.

To utilize these observations, the camera calibration process was repeated on subsets of the keypoints predicted with confidence above an optimized threshold:
\begin{itemize}
\item All keypoints
\item Only line-line intersection keypoints
\item Ground plane keypoints after filtering potential outliers using RANSAC filtering (ground plane points that could not be fitted by the homography reprojection with a 5 px tolerance were excluded)
\item All points on the ground plane 
\end{itemize}

Subsets including keypoints on the crossbars were treated with the OpenCV Multiplane algorithm, while the rest were treated with the standard OpenCV calibration method. The known dimensions of the pitch pattern enabled the use of pattern reprojection as a robust internal standard for assessing the correctness of the predicted camera parameters. The resulting camera calibration values were determined through a voting process based on the Root Mean Square Error (RMSE) of the pitch pattern reprojection. The camera parameter set resulting in the lowest RMSE was chosen as the final prediction, with preference given to the parameters based on all detected points if the RMSE value was less than 5 px (referred to as the Voter).

To further exploit the approach based on subsampling the predicted points and using subsets of reliable keypoints where possible, we implemented the Iterative Voter algorithm. This algorithm iterates over the Voter process described above, using points with predicted confidence above three threshold levels, and selects the camera parameter set obtained from the points with the highest confidence. The keypoint detection confidence thresholds were optimised using Optuna \cite{optuna_2019} to maximize the Score value on the \emph{valid} dataset.

Finally, we fused the line prediction model results: when the number of available predicted keypoints within the image was insufficient, predicted line intersection points, even beyond the image boundary, were considered. This also allowed the addition of points missed by the point model.

For all calibration algorithms, camera calibration results were discarded completely if the camera parameters were unrealistic based on the following simple heuristic thresholds. A camera was considered unreasonable if the results suggested it was below ground level (or higher than 100m above the ground), more than 250m away from the pitch center in either spatial dimension, or if the lens focal length was outside the range [10, 20000] px.

\begin{table}
    \caption{Impact of keypoint subsets, as determined using the \emph{valid} split. The keypoint subsets were: Intersections - 30 line-line intersections and 6 line-conic intersections; Tangent - 8 tangent points on the conics; Extra - 9 additional points along the pitch longitudinal axis and 4 quarter-turn points on the central circle.}
    \label{tab:abl_keypoints}
    \begin{tabular}{|ccc|cccc|}
        \toprule
        \multicolumn{3}{|c|}{Points} & \multicolumn{4}{|c|}{Metrics} \\
        Intersections & Tangent & Extra & $L_{2}$ (px) & Acc@5 & CR & Score\\
        \midrule
        + & - & - & 4.46 & 0.7711 & 0.6034 & 0.4653 \\
        + & + & - & 4.79 & 0.7527 & 0.6731 & 0.5067 \\
        + & + & + & 4.89 & 0.7446 & 0.7245 & 0.5395 \\
        \bottomrule
    \end{tabular}
\end{table}

\subsection{Ablation Study}

In the section we study the impact of the proposed keypoints and algorithmic decisions on the camera calibration pipeline. The reported results were obtained on SoccerNet 2023 dataset.

\subsubsection{Impact of the generated keypoints}

The study of the impact of keypoint subsets was performed by training the keypoint detection model with the specified subset of keypoints. The camera calibration algorithm for the stage was three-plane calibration, providing a reasonable balance between algorithm complexity and generality.

As shown in Table \ref{tab:abl_keypoints}, additional points enhance the completeness of the predictions by enabling camera calibration in more challenging situations, for instance, where there are not enough visible intersection points. Notably, the Acc@5 metric achieves the best result when only intersections were utilised. This may be attributed to the fact that intersection points are the most easily identifiable, as they are marked with clear geometric markings on the football pitch, thereby allowing for less ambiguous and more accurate neural network predictions. Conversely, tangent points and the extra points added by homography projection lack the marking and may have less accurate computed annotations due to the points generation method nature. This hypothesis is supported by the observed trend of increasing \( L_{2} \) errors with the addition of the points. Simultaneously, additional keypoints demonstrably benefit the completeness rate, facilitating camera calibration on a greater number of samples and, consequently, increasing the overall Score metric.

\subsubsection{Camera calibration algorithms}

\begin{table}
    \caption{Camera calibration algorithms comparison on the \emph{test} split of SoccerNet 2023, which was not used for training and the algorithms tuning.}
    \label{tab:abl_calibration}
    \begin{tabular}{|l|ccc|}
        \toprule
        Algorithm & Acc@5 & CR & Score\\
        \midrule
        OpenCV (reference) & 0.7303 & 0.7252 & 0.5297 \\
        OpenCV Multiplane & 0.7595 & 0.7221 & 0.5484 \\
        Voter &  0.7759 & 0.7173 & 0.5566 \\
        Iterative Voter & 0.7668 & 0.7343 & 0.5630 \\
        Iterative Voter + Lines & 0.7663 & 0.7358 & 0.5638 \\
        \bottomrule
    \end{tabular}
\end{table}

Table \ref{tab:abl_calibration} demonstrates the impact of different calibration algorithms applied to the models predictions.

Multiplane calibration resulted in substantial improvements to the accuracy metric compared to the reference calibration algorithm. This illustrates the positive impact of integrating non-planar points into the camera calibration process. The improvement can be explained by the fact that in many video frames, the crossbars are located higher than the rest of the ground plane keypoints. This vertical distribution of points improves the coverage of the frame with calibration points, leading to better calibration results.

Adaptively selecting points with the Voter algorithm allowed for even more accurate calibration, albeit at the cost of decreased completeness. This behaviour can be explained by the Voter algorithm's discarding of points predicted with confidence below a certain threshold. The selected points resulted in more accurate calibration due to RANSAC removing potential outliers and the selection of the most plausible parameters based on the reprojection RMSE.

The drawback of the Voter algorithm, which could lead to $CR$ reduction in some cases, was addressed by the Iterative Voter process. This iterative approach allows calibration to be performed on points with lower confidence if there are not enough high-confidence points available. By gradually incorporating points with lower confidence, the Iterative Voter process strikes a balance between accuracy and completeness.

Finally, integrating the lines model results further increased completeness, resulting in an overall improvement in the Score. The final algorithm, incorporating all these heuristical refinements, was used for the best submission to the SoccerNet Camera Calibration Challenge 2023.

\begin{table}
    \caption{Top-5 SoccerNet Camera Calibration Challenge 2023 entries \cite{cioppa2023soccernet}.}
    \label{tab:abl_leaderboard}
    \begin{tabular}{|cl|ccc|}
        \toprule
        Rank & Method & Acc@5 & CR & Score\\
        \midrule
        NA & Baseline & 0.1354 & 0.6154 & 0.0833 \\
        5 & ikapetan & 0.5378 & 0.7971 & 0.4287 \\
        4 & BPP & 0.6912 & 0.7254 & 0.5014 \\
        3 & SAIVA Calibration & 0.6033 & 0.8722 & 0.5262 \\
        2 & Spiideo & 0.5295 & 0.9997 & 0.5293 \\
        1 & \textbf{Our} & \textbf{0.7322} & 0.7559 & \textbf{0.5535} \\
        \bottomrule
    \end{tabular}
\end{table}

\subsection{Challenge split comparison}

Part of SoccerNet Camera Calibration 2023 dataset was kept private without publically available ground truth and used for the evaluation of the Challenge submission on eval.ai platform \cite{EvalAI}, serving as an independent verification of the entries. The top part of the leaderboard is reproduced in Table \ref{tab:abl_leaderboard}.

The baseline solution by the SoccerNet team employed two step approach with DeepLabv3 \cite{DeepLabv3} semantic segmentation of the annotated lines, followed by line extremities finding algorithms. The resulted extremities were used to get homography from the football pitch model to the image coordinates. The homography was then decomposed to get the camera parameters.

The third best solution utilised two calibration components: object detection,
trained on a homography matrix with field transformation data, which was used for sampling the keypoint detection results evaluated based on distance from segmentation lines. The keypoints detection model utilised denser 221 keypoints \cite{cioppa2023soccernet}.

The second solution by the Spiideo team was based on semantic segmentation of six classes of different parts of the football pitch, utilising DeepLabv3 \cite{DeepLabv3} with ground truth data produced by floodfill of pitch parts bounded by the annotation polylines. Predicted semantic segmentation maps and rendered a 3D football pitch image given camera parameters, were treated by differential renderer SoftRasterizer \cite{SoftRasterizer} to optimise the camera parameters.

The challenge results demonstrate that a relatively small number of strategically positioned keypoints and a comparatively simple deep learning architecture of our approach are sufficient to achieve the best accuracy with a substantial margin. The applied heuristic post-processing secured the best result based on the Score metric, effectively maintaining the trade-off between completeness and accuracy.

\section{Conclusion}
\label{sec:conclusion}

This paper presents a novel and effective multi-stage pipeline for camera calibration in soccer broadcast video frames. Our method addresses the critical challenge of finding a sufficient number of high-quality point pairs for accurate calibration by leveraging the inherent structural features of the football pitch. This includes exploiting line-line and line-conic intersections, points on the conics themselves, and other geometric features of a football pitch, thereby significantly increasing the number of usable points and enhancing both accuracy and robustness.

We address imperfect annotation using line and conic fitting, employ deep learning techniques for keypoint and line detection, and incorporate geometric constraints based on real-world pitch dimensions. Our approach fuses off-plane points and utilizes a voter algorithm to iteratively select the most reliable keypoints for calibration, further improving accuracy and optimizing the trade-off between completeness ratio and accuracy. We demonstrate the effectiveness of our pipeline through extensive experiments on the SoccerNet Camera Calibration 2023 dataset, achieving state-of-the-art performance and securing the top position in the challenge.

The results highlight the importance of integrating domain knowledge and structural insights into camera calibration pipelines. Our approach demonstrates that a relatively small number of strategically positioned keypoints, combined with a comparatively simple deep learning architecture and effective heuristic post-processing, can achieve superior results compared to more complex methods.

This pipeline opens up future research opportunities, including exploring the direct incorporation of ellipses and lines into the calibration process, rather than relying solely on keypoints. Additionally, we aim to investigate the temporal stability of predictions across frames from broadcast videos, potentially through the utilization of temporal smoothing techniques and neural networks based on frame sequence processing.

\begin{acks}

The authors would like to thank the SoccerNet Challenge 2023 organizers for providing this opportunity and the dataset. We are also grateful to EVS Broadcast Equipment for sponsoring the Camera Calibration task of the Challenge.

Additionally, the authors wish to express their sincere gratitude to their families and friends for their unwavering support and encouragement throughout the challenge.

\end{acks}



\end{document}